# PathologyBERT - Pre-trained Vs. A New Transformer Language Model for Pathology Domain

**Thiago Santos, MS[1], Amara Tariq, Ph.D.[2], Susmita Das, MS[3], Kavyasree Vayalpati, MS[4], Geoffrey H. Smith, MD[5], Hari Trivedi, MD[6], Imon Banerjee, Ph.D[2,4]**

[1]Emory University, Department of Computer Science, Atlanta, Georgia, USA; [2]Mayo Clinic, Phoenix, Arizona, USA; [3]Indian Institute of Technology (IIT), Centre of Excellence in Artificial Intelligence, Kharagpur, West Bengal, India; [4]Arizona State University, School of Computing and Augmented Intelligence, Tempe, Arizona, USA; [5]Emory University, Department of Pathology, Atlanta, Georgia, USA; [6]Emory University, Department of Radiology, Atlanta, Georgia, USA;

**Abstract** *Pathology text mining is a challenging task given the reporting variability and constant new findings in cancer sub-type definitions. However, successful text mining of a large pathology database can play a critical role to advance 'big data' cancer research like similarity-based treatment selection, case identification, prognostication, surveillance, clinical trial screening, risk stratification, and many others. While there is a growing interest in developing language models for more specific clinical domains, no pathology-specific language space exist to support the rapid data-mining development in pathology space. In literature, a few approaches fine-tuned general transformer models on specialized corpora while maintaining the original tokenizer, but in fields requiring specialized terminology, these models often fail to perform adequately. We propose PathologyBERT - a pre-trained masked language model which was trained on 347,173 histopathology specimen reports and publicly released in the Huggingface[1] repository. Our comprehensive experiments demonstrate that pre-training of transformer model on pathology corpora yields performance improvements on Natural Language Understanding (NLU) and Breast Cancer Diagnose Classification when compared to nonspecific language models.*

## 1 Introduction

Pathology reports document specific observations of tissue specimens which can play a critical role to advance 'big data' cancer research like similarity-based treatment selection, case identification, prognostication, surveillance, clinical trial screening, risk stratification, and many others[2-4]. However, due to the high variability in language and documentation templates, the content of pathology reports can be difficult to mine automatically. The complexity of pathology ontology structuring, clinical diagnoses interspersed with complex explanations, different terminology to label the same cancer, synonymous and ambiguous terms, and multiple diagnoses in a single report are just a few of such challenges. Recent advances in Natural Language Processing (NLP) techniques can be leveraged for better understanding of contextual relations in pathology text mining by exploiting attention based Encoder-Decoder[5] architectures. The Bidirectional Encoder Representations from Transformers (BERT)[6] is a contextualized language representation model based on a multilayer bi-directional encoder, where the transformer neural network uses parallel attention layers rather than sequential recurrence. Therefore, BERT model can represent words or sequences in a way that captures the contextual information, causing the same sequence of words to have different representations when they appear in different contexts. As a result, previous state-of-the-art language models such as Word2Vec[7,8], fastText[9], ELMo[10], and ULMFiT[11] were outperformed by BERT[6] in several targeted NLP tasks.

Recently, several studies have explored the utility and efficacy of contextual models in the biomedical domains[12-15]. Lee et al.[12] trained a BERT model using a corpus of biomedical research articles derived from PubMed abstracts and PubMed Central full texts to achieve new state-of-the-art performance on several text mining tasks, including named entity recognition, relation extraction, question answering. In order to provide a standardized benchmark for comparison of different models, Peng et al.[15] created the Biomedical Language Understanding Evaluation (BLUE), a benchmark consisting of five tasks (i.e., sentence similarity, named entity recognition, relation extraction, document multi-label classification and inference). In addition, the authors released BlueBERT, a BERT model trained using PubMed abstracts and MIMIC III[16] (Medical Information Mart for Intensive Care III) clinical notes. Alsentzer et al.[13] released a clinical BERT base model trained on the MIMIC III[16] database. Beltagy et al.[14] released SciBERT which is trained on a random sample of 1.14M full-text scientific papers from Semantic Scholar (18% computer science papers,

82% biomedical papers).

While there is a growing interest in developing language models for more specific clinical domains, the current trend appears to prefer fine-tuning general transformer models on specialized corpora rather than developing models from the ground up with specialized vocabulary[12,13,15]. However, in fields requiring specialized terminology, such as pathology, these models often fail to perform adequately. One of the major reasons for this limitation is because often transformerbased models employ WordPiece[17]for unsupervised input tokenization, a technique that relies on a predetermined set of Word-Pieces[17]. The word-piece vocabulary is built such that it contains the most commonly used words or sub-word units and any new words (out-of-vocabulary) can be represented by frequent subwords. Although WordPiece was built to handle suffixes and complex compound words, it often fails with domain-specific longer terms. As example, while ClinicalBERT[13]successfully tokenizes the word "endpoint" as ['end', ##point], it tokenize the word "carcinoma" as ['car', '##cin', '##oma'] in which the word lost its actual meaning and replaced by some non-relevant junk words, such as 'car'. The words which was replaced by the junk pieces, may not play the original role in deriving the contextual representation of the sentence or the paragraph, even when analyzed by the powerful transformer models. In summary, word-Piece tokenizer have three major drawbacks: (1) they do not retain the the full word representation of every word in the vocabulary, (2) they often fail to tokenize domain-specific compound words, and (3) they are brittle to noise, meaning that even minor typos may affect the representation of a word. Training a domain specific transformer model is often challenging as it requires a large amount of domain specific training data. Lack of availability of the biomedical domain specific data is a constraint for training the BERT model for many biomedical NLP tasks.

In this work, we propose PathologyBERT - a pre-trained masked language model which was trained on 347,173 histopathology specimen reports. As far as we know, we are the first authors to propose and publicly release a pretrained masked language model for pathology domain. The contributions of the study can be summarized as following: *i*) We train and publicly release PathologyBERT - a transformer-based model trained on Histopathology specimens text; *ii*) We perform extensive experimentation and comparison between PathologyBERT and other publicly available BERT models to investigate the performance of masked word prediction; and *iii*) finally, compared a targeted task performance between PathologyBERT and other publicly available BERT models to show the utility of the language space.

The rest of this paper is organized as follows. Section 2 presents the details of the methodology, including the proposed BERT model architecture, the training processes, and detailed analysis of the dataset. Section 3 describes the experiment design and analyzes the results of experiments. Section 4 concludes the paper by describing the limitations and setting forth future work.

## 2 Methods

Figure 1 presents the core steps of the proposed pipeline by highlighting different training steps. First, we employed experts to generate an annotated dataset with diagnosis labels from randomly selected samples. Second, we used the unsupervised dataset to train and evaluate the masked language model. Finally, both supervised and unsupervised datasets are used to assess the masked language prediction and classification performance on the annotated data. Each processing block (gray background) is described in the following section.

### 2.1 Dataset

In order to train the model and validate the language space, we collected the following two non-overlapping corpora of histopathology reports from Emory University Hospital (EUH).

**Corpus I. Unlabeled pathology reports from EUH** - With the approval of Emory University Institutional Review Board(IRB), a total of 340,492 unstructured Histopathology specimens from 67,136 patients were extracted from the clinical data warehouse of Emory University Hospital (EUH) between the years of 1981 and 2021. These reports are written following a semi-structured template, and contain standard sections such as: "HISTORY", "DIAGNOSIS","GROSS DESCRIPTION", "MICROSCOPY EXAMINATION", and "COMMENTS". Although all sections may contain relevant pathology information, due to the size constraint of the BERT input sequence[6], we only incorporated the "DIAGNOSIS" section in this study. Using simple regex, we divided each part from a single specimen into an individual diagnosis element. The average report length is $42 \pm 26$ tokens resulting in a corpus size of approximately 7.2M tokens.

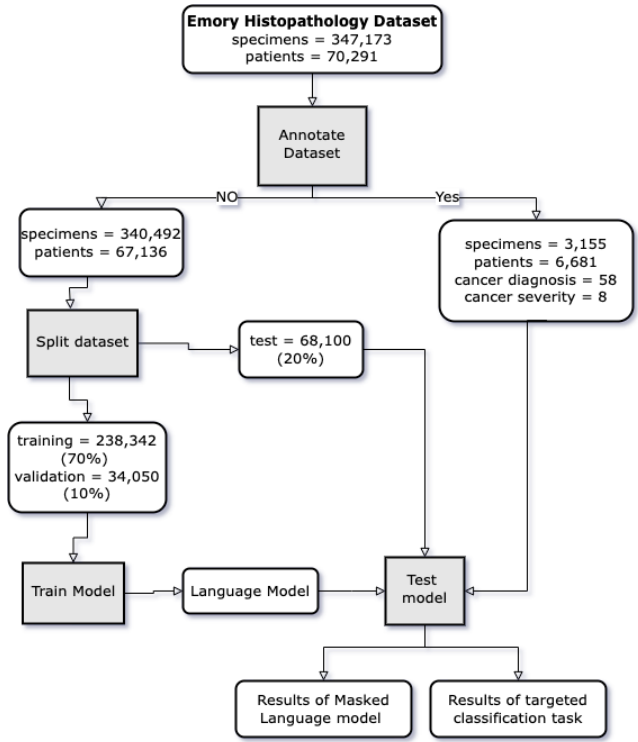

**Figure 1:** Flowchart diagram of the proposed pipeline - training and evaluation. Gray background square boxes highlight the processing components.

The reports may contain varying year-wise templates which makes the language space generation very challenging. The quality of extracted text features has been demonstrated to be improved by data pre-processing[18]. When building the language model vocabulary, we employ pre-processing to minimize the feature space and improve data generalization. In this work, we transformed the segmented data through a series of standard pre-processing techniques. This was accomplished by applying a conversion to lowercase and removal of numbers. In addition, we discarded identifying details of pathologist, clinicians, and patients from the reports. A descriptive statistics across years are illustrated on Table 1. We train the masked language PathologyBERT model using only Corpus 1. We leveraged corpus I to train, test and validate the masked language model by randomly splitting the corpus into 238,342 (70%) for training, 34,050 (10%) for validation, and 68,100 (20%) for testing purpose.

**Corpus II. Manually labeled pathology reports from EUH** - In order to evaluate the influence of vocabulary coverage on the common masked language model prediction and domain-specific breast cancer diagnose task, we have collected a total of 6,681 unstructured histopathology specimens from 3,155 patients from the EUH clinical data warehouse between the years of 2010 and 2021. This corpus does not overlap with corpus I as there is no intersection of patients between corpus. Each part was classified into fifty eight benign and malignant pathologic diagnoses (invasive ductal carcinoma, lobular carcinoma in situ, radial scar, fibroadenoma, etc.) and grouped into six categories (invasive breast cancer, in situ breast cancer, high risk lesion, non-breast cancer, benign, and negative.). Each part was labeled with at least one label but no upper limit on the number of possible labels. The whole corpus was used in our study to evaluate the model performance.

### 2.2 Pathology Language Space Training

We trained the traditional 12-layers transformer block architecture of BERT starting from random weights at each block. Rather than demonstrating technical novelty in the model design, the focus of this research is to generate pre-trained embeddings of pathology language space for open-source model development and improve performance on the following targeted tasks. Hence our training procedure follows the original BERT-Base implementation[6] while trained on histopathology specimens text. For all pre-training experiments, we leverage the PyTorch[19] HuggingFace[1] implementation of BERT and WordPiece tokenizer[17].

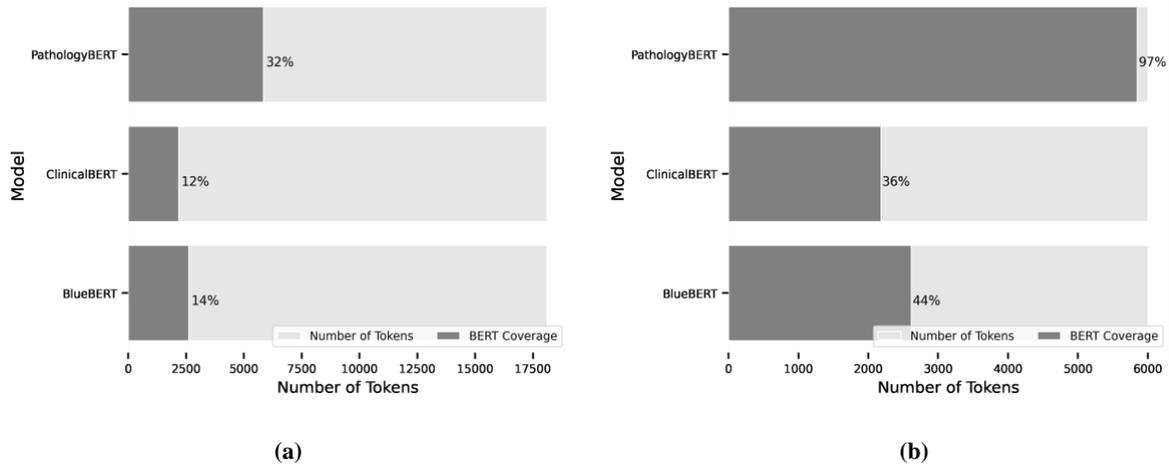

**Figure 2:** WordPiece tokenizer full word coverage over Corpus I for: (a) All unique tokens; (b) Only tokens with minimal frequency of 10.

| Dataset Statistics | Year of reporting | | | | | | | | | | | |
|---|---|---|---|---|---|---|---|---|---|---|---|---|
| | 1981-1984 | 1985-1988 | 1989-1992 | 1993-1996 | 1997-2000 | 2001-2004 | 2005-2008 | 2009-2012 | 2013-2015 | 2016-2019 | 2020-2021 | Overall |
| No. of patients | 70 | 300 | 227 | 1,507 | 5,716 | 8,150 | 8,429 | 10,061 | 11,789 | 18,584 | 2,303 | 67,136 |
| Number of reports | 112 | 1,020 | 1,247 | 4,172 | 12,950 | 26,346 | 36,777 | 53,369 | 66,599 | 115,775 | 22,125 | 340,492 |
| Mean report size | 10±2 | 30±21 | 40±50 | 45±36 | 30±28 | 44±22 | 48±29 | 42±32 | 25±11 | 39±20 | 32±23 | 42±26 |
| No. of words | 837 | 20,381 | 31,691 | 111,651 | 319,387 | 950,367 | 1,253,935 | 1,263,248 | 1,024,134 | 1,884,896 | 365,445 | 7,225,972 |
| No. of tokens | 139 | 1,386 | 2,067 | 3,231 | 4,093 | 7,068 | 7,586 | 7,396 | 8,115 | 9,529 | 5,618 | 18,355 |

**Table 1:** Descriptive statistics from corpus 1 across years.

**Vocabulary -** BERT uses WordPiece[17] for unsupervised input tokenization. The vocabulary is built such that it contains the most commonly used words or sub-words. As a result, any new words can be represented by frequent sub-words (e.g. BERT tokenizes dysplasia as d ##ys ##p ##lasia). We trained a new WordPiece vocabulary on our pathology Corpus I in order to create a new PathologyBERT vocabulary. In this study, we only produced uncased vocabularies, and the vocabulary size was set to 13K after an extensive hyperparameter search. The resulting token coverage comparison between ClinicalBERT[13], BlueBERT[15], and PathologyBERT is illustrated in Fig 2. The findings show a substantial difference in frequently used words between pathology and other medical domain texts.

**Pre-training -** We explored the following hyperparameters: learning rate $\in \{2e-4, 2e-5\}$, batch size $\in \{32, 64\}$, word minimal frequency $\in \{2, 4\}$, and vocabulary size $\in \{10k, 13k, 15k\}$. The best model was trained with a batch size of 32, a maximum sequence length of 64 (mean size of report is 42 ± 26), a learning rate of 2e-5, and masked language model probability of 0.15. Models were trained for 300,000 steps. All other BERT default parameters were used. We used a GeForce Quadro RTX 6000 24 GB GPUs. The extensive hyperparameter searching on our corpus I takes approximately 5 days when using multiple available GPUs. Finally, training the PathologyBERT model with optimal hyperparameter from scratch on our corpus 1 takes 36 hours.

### 2.3 Targeted tasks

With minimal architectural modification, PathologyBERT can be applied to various downstream text mining tasks. Within the scope of this study, we applied the learnt language space for two targeted tasks - (i) masked language prediction with different thresholds to show the semantic understanding of the model; and (ii) classification of reports according to cancer severity.

| Diagnose | PathologyBERT | | ClinicalBERT | | BlueBERT | | Support |
|---|---|---|---|---|---|---|---|
| | Accuracy | Top 5 Accuracy | Accuracy | Top 5 Accuracy | Accuracy | Top 5 Accuracy | |
| Apocrine Carcinoma | 0.64 | 0.89 | 0.38 | 0.5 | 0.5 | 0.56 | 32 |
| Apocrine Metaplasia | 0.67 | 0.85 | 0.26 | 0.49 | 0.39 | 0.56 | 4724 |
| Atypical Ductal Hyperplasia | 0.68 | 0.75 | 0.25 | 0.47 | 0.29 | 0.52 | 2661 |
| Atypical Lobular Hyperplasia | 0.66 | 0.82 | 0.26 | 0.42 | 0.36 | 0.54 | 1704 |
| Atypical Papilloma | 0.62 | 0.81 | 0.28 | 0.41 | 0.34 | 0.48 | 337 |
| Atypical Phyllodes | 0.68 | 0.8 | 0.39 | 0.61 | 0.52 | 0.65 | 23 |
| Biopsy Site Changes | 0.60 | 0.74 | 0.30 | 0.44 | 0.30 | 0.52 | 7686 |
| Columnar Cell Change with Atypia | 0.66 | 0.82 | 0.26 | 0.39 | 0.30 | 0.5 | 513 |
| Columnar Cell Change without Atypia | 0.68 | 0.88 | 0.32 | 0.46 | 0.34 | 0.52 | 3436 |
| Cyst | 0.7 | 0.80 | 0.27 | 0.5 | 0.44 | 0.60 | 2730 |
| Ductal Carcinoma in Situ | 0.60 | 0.86 | 0.33 | 0.48 | 0.42 | 0.56 | 12003 |
| Excisional or Post-surgical Change | 0.68 | 0.81 | 0.28 | 0.5 | 0.40 | 0.57 | 231 |
| Fat Necrosis | 0.7 | 0.82 | 0.25 | 0.44 | 0.42 | 0.59 | 1353 |
| Fibroadenoma | 0.70 | 0.85 | 0.24 | 0.47 | 0.42 | 0.59 | 3265 |
| Fibroadenomatoid | 0.66 | 0.84 | 0.3 | 0.46 | 0.39 | 0.54 | 713 |
| Fibrocystic Disease | 0.63 | 0.87 | 0.24 | 0.42 | 0.36 | 0.55 | 7027 |
| Fibromatoses | 0.63 | 0.77 | 0.2 | 0.35 | 0.36 | 0.52 | 85 |
| Fibrosis | 0.68 | 0.84 | 0.3 | 0.46 | 0.38 | 0.54 | 545 |
| Flat Epithelial Atypia | 0.62 | 0.8 | 0.34 | 0.55 | 0.38 | 0.57 | 1034 |
| Fna - Benign | 0.73 | 0.89 | 0.44 | 0.59 | 0.56 | 0.7 | 216 |
| Fna - Malignant | 0.65 | 0.87 | 0.35 | 0.5 | 0.46 | 0.59 | 166 |
| Grade I | 0.65 | 0.82 | 0.32 | 0.52 | 0.37 | 0.56 | 2905 |
| Grade II | 0.6 | 0.86 | 0.32 | 0.49 | 0.43 | 0.56 | 4352 |
| Grade III | 0.64 | 0.88 | 0.27 | 0.46 | 0.38 | 0.57 | 2169 |
| Granular Cell Tumor | 0.63 | 0.81 | 0.3 | 0.44 | 0.48 | 0.67 | 27 |
| Hamartoma | 0.66 | 0.83 | 0.25 | 0.67 | 0.5 | 0.67 | 12 |
| Hemangioma | 0.7 | 0.84 | 0.21 | 0.36 | 0.33 | 0.51 | 105 |
| High | 0.64 | 0.83 | 0.3 | 0.48 | 0.4 | 0.62 | 3813 |
| Hyperplasia with Atypia | 0.67 | 0.87 | 0.28 | 0.49 | 0.38 | 0.56 | 39 |
| Intermediate | 0.68 | 0.86 | 0.28 | 0.52 | 0.37 | 0.55 | 4737 |
| Intracystic Papillary Carcinoma | 0.68 | 0.72 | 0.29 | 0.46 | 0.32 | 0.50 | 102 |
| Intraductal Papillary Carcinoma | 0.64 | 0.76 | 0.32 | 0.48 | 0.38 | 0.54 | 296 |
| Intraductal Papilloma | 0.67 | 0.82 | 0.24 | 0.36 | 0.3 | 0.48 | 4355 |
| Invasive Ductal Carcinoma | 0.62 | 0.87 | 0.36 | 0.5 | 0.42 | 0.56 | 9920 |
| Invasive Lobular Carcinoma | 0.62 | 0.78 | 0.31 | 0.50 | 0.42 | 0.56 | 1181 |
| Lactational Change | 0.7 | 0.74 | 0.30 | 0.49 | 0.38 | 0.55 | 174 |
| Lobular Carcinoma in Situ | 0.66 | 0.79 | 0.28 | 0.46 | 0.4 | 0.54 | 1779 |
| Low | 0.66 | 0.75 | 0.4 | 0.56 | 0.45 | 0.57 | 2222 |
| Lymph Node - Benign | 0.72 | 0.87 | 0.25 | 0.45 | 0.33 | 0.53 | 3449 |
| Lymph Node - Metastatic | 0.76 | 0.9 | 0.25 | 0.41 | 0.38 | 0.55 | 1618 |
| Lymphoma | 0.67 | 0.81 | 0.29 | 0.48 | 0.44 | 0.57 | 141 |
| Malignant (Sarcomas) | 0.72 | 0.88 | 0.32 | 0.48 | 0.39 | 0.52 | 114 |
| Medullary Carcinoma | 0.67 | 0.88 | 0.25 | 0.38 | 0.33 | 0.62 | 24 |
| Metaplastic Carcinoma | 0.64 | 0.89 | 0.27 | 0.50 | 0.38 | 0.55 | 197 |
| Microscopic Papilloma | 0.65 | 0.78 | 0.26 | 0.46 | 0.34 | 0.53 | 519 |
| Mucinous Carcinoma | 0.61 | 0.89 | 0.32 | 0.48 | 0.35 | 0.55 | 312 |
| Mucocele | 0.62 | 0.71 | 0.3 | 0.45 | 0.27 | 0.47 | 135 |
| Negative | 0.70 | 0.88 | 0.33 | 0.47 | 0.41 | 0.68 | 4433 |
| Non-breast Metastasis | 0.61 | 0.70 | 0.44 | 0.62 | 0.37 | 0.51 | 68 |
| Pagets | 0.69 | 0.81 | 0.35 | 0.52 | 0.37 | 0.53 | 97 |
| Phyllodes | 0.72 | 0.84 | 0.39 | 0.61 | 0.44 | 0.64 | 285 |
| Pseudoangiomatous Stromal Hyperplasia | 0.72 | 0.86 | 0.22 | 0.39 | 0.29 | 0.5 | 236 |
| Radial Scar | 0.65 | 0.81 | 0.28 | 0.44 | 0.28 | 0.5 | 865 |
| Sclerosing Adenosis | 0.64 | 0.82 | 0.25 | 0.42 | 0.32 | 0.5 | 3057 |
| Seroma | 0.71 | 0.83 | 0.3 | 0.45 | 0.49 | 0.62 | 76 |
| Tubular Carcinoma | 0.6 | 0.79 | 0.34 | 0.53 | 0.32 | 0.5 | 524 |
| Uncertain | 0.71 | 0.82 | 0.29 | 0.41 | 0.38 | 0.54 | 56 |
| Usual Ductal Hyperplasia | 0.64 | 0.82 | 0.28 | 0.46 | 0.32 | 0.55 | 6646 |
| **Mean** | **0.66** | **0.82** | 0.29 | 0.47 | 0.38 | 0.55 | 111524 |

**Table 2:** Inference table results across diagnosis from masked language prediction over corpus II. We randomly mask 15% of tokens per diagnose report, resulting in a total of 111,524 samples. In addition to Accuracy, we also report the Top 5 Accuracy, which takes the top five predictions into account. We consider a prediction correct if it matches one of the top five predictions.

| Model | Masked 15% | | Masked 30% | | Masked 45% | | Masked 60% | | Masked 75% | |
|---|---|---|---|---|---|---|---|---|---|---|
| | Accuracy | Top 5 Accuracy | Accuracy | Top 5 Accuracy | Accuracy | Top 5 Accuracy | Accuracy | Top 5 Accuracy | Accuracy | Top 5 Accuracy |
| ClinicalBERT | 0.26 | 0.43 | 0.26 | 0.42 | 0.25 | 0.42 | 0.24 | 0.4 | 0.23 | 0.39 |
| BlueBERT | 0.37 | 0.53 | 0.36 | 0.52 | 0.35 | 0.51 | 0.33 | 0.5 | 0.31 | 0.5 |
| **PathologyBERT** | **0.73** | **0.83** | **0.75** | **0.84** | **0.74** | **0.84** | **0.74** | **0.84** | **0.74** | **0.84** |

**Table 3:** Masked language prediction results over Corpus I holdout test set with varying thresholds. We also report the Top 5 Accuracy, which takes the top five predictions into account. We consider a prediction correct if it matches one of the top five predictions.

We fine-tune PathologyBERT on Corpus II to predict 6 breast cancer diagnose severity (invasive breast cancer, high risk lesion, borderline lesion, non-breast cancer, benign, and negative). Note that each specimen has at least one label but no upper limit on the number of possible labels, making the task a multi-label prediction problem. To overcome this challenge, we reformulated each label learning as a binary classification task and transformed the multilabel learning into multiple binary classification tasks.

The architecture, optimization, and hyperparameters are essentially the same as that of presented on BERT. During the fine-tuning process, we feed the final BERT vector for the [CLS] token into a linear classification layer to predict multi-labels for each specimen report. We optimize binary cross entropy loss using Adam$_{20}$ and applied a dropout of 0.2. We fine-tune for 6 epochs using a batch size of 16, 32, or 64 and a learning rate of 5e-6, 1e-5, 2e-5, or 5e-5 with early stopping on the development set (patience of 10). The best model was trained with a batch size of 32 and a learning rate of 2e-5.

## 3 Experiments and Results

### 3.13.1 Vocabulary coverage and masked language prediction

Our experiments are focused on two aspects; 1) pathology-specific vocabulary coverage, 2) performance of masked language model. We report results on both Corpus I and Corpus II. As described in the dataset section, reports in Corpus II are labelled by diagnoses. Hence, this corpus allowed us to judge the performance of PathologyBERT stratified by the diagnosis label.

Major advantage of training a tokenizer and a transformer based language model for pathology reports is coverage of pathology-specific terminology. Figure 2 shows the overall coverage of pathology vocabulary by the proposed PathologyBERT compared to pre-trained models like ClinicalBERT and BlueBERT. It is clear that PathologyBERT provides far better coverage than pre-trained models. Pathology-BERT nearly covers all terms with a frequency of at least 10. This is intuitive as PatholgoyBERT has specifically been designed and trained on pathology reports while none of the pre-trained BERT models had this opportunity.

On the unseen cohort II, we also validated the coverage of the ClinicalBERT, BlueBERT and PathologyBERT for both common and extremely rare diagnosis classes. Figure 3 shows the coverage ratio of models on the unseen cohort II along with number of representative class samples. Coverage ratio is defined as $N(w_p)/N(w_n)$ where $N(w_p)$ is the number of unique words within the class member present in vocabulary and $N(w_n)$ is the total number of unique words within the class member. As seen from the figure, PathologyBERT achieved good coverage ($> 0.75$) for extremely rare classes, such as Apocrine Carcinoma (3 samples), Atypical Phyllodes (3 samples). Other transformers vocabulary coverage have only been limited to 0.5 range.

Given better vocabulary coverage, we expect to see better masked language prediction accuracy from the masked language model built on top of PathologyBERT. Results in Table 3 confirm this expectation on the hold-out validation data. We report both the accuracy and the 'Top 5 Accuracy' which counts occurrence of the masked token in the top-5 predicted tokens as a correct prediction. PathologyBERT outperforms both BlueBERT and ClinicalBERT by a large margin, i.e., accuracy of 0.73 achieved by PathologyBERT compared against 0.37 for BlueBERT and 0.26 for ClinicalBERT at conventional 15% masking rate. We extended these experiments across a wide range of masking rate, i.e., from 15% to 75%. It is interesting to note that performance remains stable, especially in terms of accuracy values, across such a wide range of masking rate for PathologyBERT. ClinicalBERT and BlueBERT drop their performances by 3 and 6 percentage points, respectively. We attribute this phenomenon to narrow frequency range of unique tokens. Only about 6000 tokens have frequency 10 or more. We believe smaller token set size results in more frequent repetition of tokens, and hence more predictable token patterns resulting in stable prediction performance even with very high masking rate. ClinicalBERT and BlueBERT fail to predict even this smaller variety of tokens because their tokenizers do not properly cover tokens included in this small but specialized token set containing pathology-specific terminology.

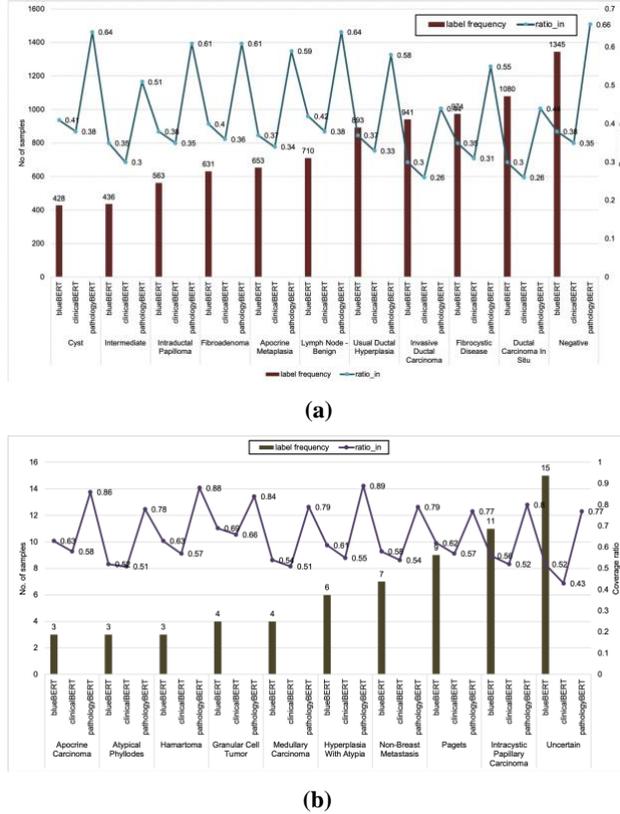

**Figure 3:** Comparative vocabulary coverage of blueBERT, clinicalBERT and pathology BERT: (a) top majority classes, and (b) minority classes.

Table 2 presents the masked language prediction results (masked 15%) for each class present in the unseen Corpus II. Similar test observation (Table 3) holds for the unseen dataset - PathologyBERT provides better coverage than pretrained BERT models on all diagnoses. Overall prediction accuracy is much higher in PathologBERT (71% accuracy) compared to ClinicalBERT (27%) and BlueBERT (38%). Interestingly, even for the rare classes with <20 occurrences, PathologyBERT obtained decent accuracy while other models achieved very low performance with both strict and relaxed metrics.

In order to present the concept more intuitively, we show in Table 4 a few randomly chosen examples of prediction by masked language models. It is interesting to note that PathologyBERT not only makes more accurate predictions, its top few suggestions consist of words pointing to similar concept or synonyms. This indicates model's ability in understanding the context of the word and its suggesting potential tokens suitable to the same context. When PathologyBERT predicts 'mastectomy' as top candidate, it also predicts 'resection' and 'excision' as second and third choices. All of these words point to the idea of tissue removal. Similarly, when 'calcification' is predicted as the top word, the second choice is 'micro-calcification'.

### 3.2 Breast Cancer Diagnose Severity Classification

Language models are widely used for generating semantically rich embeddings for words or word sequences that can be further used for downstream prediction tasks. Performance of downstream predictor is another indicator of the quality of the embedding generated by language model, and hence the language model itself. We evaluated the proposed language model PathologyBERT for downstream prediction of breast cancer diagnosis on Corpus II. As illustrated in Table 5, PathologyBERT based predictor outperforms predictors based on BlueBERT, and ClinicalBERT. It is important to note that the largest performance improvement is achieved for Non-breast cancer (NBC) label (0.59 f1-score of BlueBERT to 0.70 f1-score of PathologyBERT). This label has the lowest support value, indicating it to be the rarest of all considered labels. It is an intuitive assumption that tokens indicative of a rare label are also rare and are not covered by BlueBERT and ClinicalBERT, but are covered by the tokenizer of PathologyBERT. Thus PathologyBERT outperforms other BERT based predictors by a significant margin for this label.

| Sentence | Masked Language Model | | | | | |
|---|---|---|---|---|---|---|
| | PathologyBERT | | ClinicalBERT | | BlueBERT | |
| | Mask Prediction | Confdence | Mask Prediction | Confdence | Mask Prediction | Confdence |
| right breast, core needle biopsy: [ **invasive** ] ductal carcinoma, nottingham gradethree, with a lobular growth pattern | **invasive** | **0.96** | invasive | 0.8 | invasive | 0.99 |
| | infltrating | 0.03 | small | 0.02 | a | 0.002 |
| | microinvasive | 0.001 | a | 0.007 | left | 0.001 |
| usual ductal epithelial hyperplasia, apocrine [**metaplasia**], calcifcations associated with benign | **metaplasia** | **0.75** | ##s | 0.26 | proliferation | 0.55 |
| | cysts | 0.13 | ##mia | 0.04 | cells | 25 |
| | adenosis | 0.1 | ##lasia | 0.09 | change | 0.1 |
| chronic [**gastritis**] with regenerative changes. no helicobacter pylori | **gastritis** | **0.79** | ##ity | 0.42 | infammation | 0.64 |
| | infammation | 0.15 | infammation | 0.16 | changes | 0.12 |
| | mucosa | 0.04 | changes | 0.12 | disease | 0.07 |
| benign fatty breast tissue including millimeters nodular fbroepithelial lesion, nodular [**adenosis**] vs fbroadenomatoid nodule | **adenosis** | **0.85** | focus | 0.69 | ##ity | 0.52 |
| | hyperplasia | 0.1 | mass | 0.11 | fat | 0.16 |
| | cyst | 0.02 | ##ity | 0.06 | scar | 0.12 |
| wire-directed segmental mastectomy: fat epithelial atypia with [ **calcifcation** ]. | **calcifcation** | **0.6** | radiation | 0.06 | invasion | 0.1 |
| | microcalcifcations | 0.36 | resolution | 0.06 | clusters | 0.06 |
| | atypia | 0.01 | improvement | 0.03 | debris | 0.05 |
| wire-directed segmental [ **mastectomy** ]: fat epithelial atypia with calcifcation. | **mastectomy** | **0.98** | study | 0.32 | sampling | 0.31 |
| | resection | 0.007 | view | 0.16 | brushing | 0.06 |
| | excision | 0.003 | assessment | 0.04 | block | 0.03 |
| intraductal papilloma with [ **sclerosis** ] and micro calcifcations with Usual ductal hyperplasia and no atypical lesions identifed. | **sclerosis** | **0.83** | focal | 0.19 | macro | 0.53 |
| | calcifcations | 0.08 | **micro** | 0.06 | central | 0.06 |
| | hyalinization | 0.009 | nuclear | 0.05 | coarse | 0.05 |
| intraductal papilloma with sclerosis and micro calcifcations with Usual ductal hyperplasia and no [ **atypical** ] lesions identifed. | **atypical** | **0.87** | focal | 0.31 | focal | 0.2 |
| | proliferative | 0.07 | other | 0.23 | mass | 0.17 |
| | invasive | 0.07 | discrete | 0.11 | other | 0.16 |
| intraductal papilloma with sclerosis and micro calcifcations with Usual ductal hyperplasia and [ **no** ] atypical lesions identifed. | **no** | **0.99** | other | 0.55 | no | 0.94 |
| | focal | 0.002 | multiple | 0.04 | some | 0.005 |
| | rare | 0.001 | no | 0.02 | multiple | 0.004 |

**Table 4:** Inference examples from masked language model. Samples are selected from holdout test set and a random token is masked. The sentence is then fed into the model for mask prediction.

| Severity category | BlueBERT | | | ClinicalBERT | | | PathologyBERT | | | Support |
|---|---|---|---|---|---|---|---|---|---|---|
| | Precision | Recall | F1 Score | Precision | Recall | F1 Score | Precision | Recall | F1 Score | |
| Invasive breast cancer-IBC | 0.98 ± 0.005 | 0.96 ± 0.004 | 0.97 ± 0.004 | 0.98 ± 0.004 | 0.98 ± 0.008 | 0.98 ± 0.004 | **0.98 ± 0.009** | **0.98 ± 0.007** | **0.98 ± 0.008** | 258 |
| Non-breast cancer-NBC | 0.89 ± 0.28 | 0.44 ± 0.2 | 0.59 ± 0.18 | 0.72 ± 0.27 | 0.31 ± 0.14 | 0.38 ± 0.17 | **0.79 ± 0.09** | **0.67 ± 0.11** | **0.70 ± 0.17** | 11 |
| In situ breast cancer-ISC | 0.96 ± 0.006 | 0.93 ± 0.005 | 0.94 ± 0.005 | 0.98 ± 0.006 | 0.96 ± 0.01 | 0.97 ± 0.007 | **0.98 ± 0.008** | **0.98 ± 0.05** | **0.98 ± 0.004** | 220 |
| High risk lesion-HRL | 0.92 ± 0.004 | 0.90 ± 0.005 | 0.91 ± 0.004 | 0.97 ± 0.01 | 0.95 ± 0.01 | 0.96 ± 0.008 | **0.97 ± 0.02** | **0.97 ± 0.03** | **0.97 ± 0.009** | 248 |
| Benign-B | 0.92 ± 0.006 | 0.96 ± 0.006 | 0.94 ± 0.006 | 0.97 ± 0.005 | 0.98 ± 0.003 | 0.98 ± 0.003 | **0.98 ± 0.01** | **0.98 ± 0.002** | **0.98 ± 0.002** | 749 |
| Negative | 0.92 ± 0.007 | 0.89 ± 0.02 | 0.90 ± 0.005 | 0.98 ± 0.008 | 0.96 ± 0.01 | 0.97 ± 0.007 | **0.98 ± 0.01** | **0.98 ± 0.02** | **0.98 ± 0.009** | 269 |
| | | | | | | | | | | |
| Accuracy | | 0.87 ± 0.007 | | | 0.92 ± 0.006 | | | **0.95 ± 0.007** | | 1755 |
| Micro Average | 0.93 ± 0.006 | 0.93 ± 0.004 | 0.93 ± 0.004 | 0.97 ± 0.004 | 0.96 ± 0.004 | 0.97 ± 0.003 | **0.98 ± 0.002** | **0.97 ± 0.003** | **0.98 ± 0.002** | |

**Table 5:** Inference table performances of BERT, BioBERT, ClinicalBERT, BlueBERT, and PathologyBERT on predicting Breast Cancer Diagnoses. Bold indicates the state-of-the-art results with 95% bootstrap confidence interval.

## 4 Conclusion

In this article, we introduced PathologyBERT, which is a pre-trained language representation model for pathology text mining. In our comprehensive experimental setting, we showed that pre-training transformer model on pathology corpora is crucial in applying it to Pathology Natural Language Understanding (NLU), information extraction, text classification, and several other text mining tasks. Further, with minimal task-specific architectural modification and with a multi-label unbalanced dataset, PathologyBERT outperforms previous models on breast cancer severity classification task. To support pathology informatics development, we will release the model with open-source license in the HuggingFace model repository. While PathologyBERT shows significant promises in regards to pathology text mining, it faces several limitations. One major limitation is that it has been trained on pathology reports collected from a single institute. Only the DIAGNOSIS section of each report was included in the training process. Different institute may use somewhat different vocabulary as well as report structure. However, given the semi-structured nature of reporting, template structure may not pose significant generalizability challenge for our model and can be fixed easily by pre-processing and section segmentation. Another limitation is smaller input size of 64 tokens as compared to 512 for BERT and 128 for ClinicalBERT. This parameter value works well with our current training corpus where we can easily extract DIAGNOSIS section with mean size of 42 tokens. However, this may pose problems when applying our model to a longer pathology reports coming from other institutions. In future, we intend to alleviate above-mentioned limitations by employing multi-institutional data for fine-tuning of our model. The scope of this work is limited to quality evaluation of the language model for its inherent task of masked token prediction and breast cancer diagnosis classification. In future, we intend to experiment with a wider variety of downstream prediction labels including named-entity recognition and hierarchical modeling for subcategories of diagnoses.

**References**

[1] Wolf T, Debut L, Sanh V, Chaumond J, Delangue C, Moi A, et al.. HuggingFace's Transformers: State-of-the-art Natural Language Processing; 2020.

[2] Schroeck F, Patterson O. Development of a Natural Language Processing Engine to Generate Bladder Cancer Pathology Data for Health Services Research. Urology vol 110. 2017:84–91.

[3] Schroeck FR, Lynch KE, Chang Jw, MacKenzie TA, Seigne JD, Robertson DJ, et al. Extent of Risk-Aligned Surveillance for Cancer Recurrence Among Patients With Early-Stage Bladder Cancer. JAMA Network Open. 2018 09;1(5):e183442–e183442.

[4] Lee J, Song H, Yoon E. Automated extraction of Biomarker information from pathology reports. BMC Medical Informatics and Decision Making. 2018.

[5] Vaswani A, Shazeer N, Parmar N, Uszkoreit J, Jones L, Gomez A, et al. Attention is all you need. 2017 06.

[6] Devlin J, Chang MW, Lee K, Toutanova K. BERT: pre-training of deep bidirectional transformers for language understanding. In: Proceedings of the 2019 Conference of the North American Chapter of the Association for Computational Linguistics). Association for Computational Linguistics; 2019. p. 4171–4186.

[7] Mikolov T, Sutskever I, Chen K, Corrado G, Dean J. Distributed representations of words and phrases and their compositionality. arXiv preprint arXiv:13104546. 2013.

[8] Mikolov T, Chen K, Corrado G, Dean J. Efficient Estimation of Word Representations in Vector Space. In: ICLR; 2013. .

[9] Bojanowski P, Grave E, Joulin A, Mikolov T. Enriching Word Vectors with Subword Information. arXiv preprint arXiv:160704606. 2016.

[10] Peters ME, Neumann M, Iyyer M, Gardner M, Clark C, Lee K, et al. Deep Contextualized Word Representations. In: Proceedings of the 2018 Conference of the North American Chapter of the Association for Computational Linguistics: Human Language Technologies, Volume 1 (Long Papers). New Orleans, Louisiana: Association for Computational Linguistics; 2018. p. 2227–2237. Available from: https://aclanthology.org/N18-1202.

[11] Howard J, Ruder S. Universal Language Model Fine-tuning for Text Classification. In: Proceedings of the 56th Annual Meeting of the Association for Computational Linguistics (Volume 1: Long Papers). Melbourne, Australia: Association for Computational Linguistics; 2018. p. 328–339. Available from: https://aclanthology.org/P18-1031.

[12] Lee J, Yoon W, Kim S, Kim D, Kim S, So C, et al. BioBERT: a pre-trained biomedical language representation model for biomedical text mining. Bioinformatics (Oxford, England). 2019 09;36.

[13] Alsentzer E, Murphy J, Boag W, Weng WH, Jindi D, Naumann T, et al. Publicly Available Clinical BERT Embeddings. In: Proceedings of the 2nd Clinical Natural Language Processing Workshop. Minneapolis, Minnesota, USA: Association for Computational Linguistics; 2019. p. 72–78. Available from: https://aclanthology.org/W19-1909.

[14] Beltagy I, Lo K, Cohan A. SciBERT: A Pretrained Language Model for Scientific Text. In: Proceedings of the 2019 Conference on Empirical Methods in Natural Language Processing and the 9th International Joint Conference on Natural Language Processing (EMNLP-IJCNLP). Hong Kong, China: Association for Computational Linguistics; 2019. p. 3615–3620. Available from: https://aclanthology.org/D19-1371.

[15] Peng Y, Yan S, lu Z. Transfer Learning in Biomedical Natural Language Processing: An Evaluation of BERT and ELMo on Ten Benchmarking Datasets; 2019. p. 58–65.

[16] Johnson A, Pollard T, Shen L, Lehman Lw, Feng M, Ghassemi M, et al. MIMIC-III, a freely accessible critical care database. Scientific Data. 2016 05;3:160035.

[17] Wu Y, Schuster M, Chen Z, Le QV, Norouzi M, Macherey W, et al. Google's Neural Machine Translation System: Bridging the Gap between Human and Machine Translation. ArXiv. 2016;abs/1609.08144.

[18] Uysal AK, Gunal S. The impact of preprocessing on text classification. Information Processing Management. 2014 01;50:104 – 112.

[19] Paszke A, Gross S, Massa F, Lerer A, Bradbury J, Chanan G, et al. PyTorch: An Imperative Style, High-Performance Deep Learning Library. In: Advances in Neural Information Processing Systems 32. Curran As sociates, Inc.; 2019. p. 8024–8035. Available from: http://papers.neurips.cc/paper/9015-pytorch-an-imperative style-high-performance-deep-learning-library.pdf.

[20] Kingma DP, Ba J. Adam: A Method for Stochastic Optimization; 2017.